\pdfoutput=1

\documentclass[11pt]{article}

\usepackage{ACL2023}

\usepackage{times}
\usepackage{latexsym}

\usepackage[T1]{fontenc}

\usepackage[utf8]{inputenc}

\usepackage{microtype}

\usepackage{inconsolata}

\usepackage{tabularx}
\usepackage{graphicx}

\title{Towards Generating Automatic Anaphora Annotations}

 \author{Dima Taji \and Daniel Zeman \\
         Charles University\\
         Faculty of Mathematics and Physics\\
         Institute of Formal and Applied Linguistics ({\'U}FAL)\\
         Prague, Czechia \\
         {\tt \{taji, zeman\}@ufal.mff.cuni.cz}}

\begin{document}
\maketitle
\begin{abstract}

Training models that can perform well on various NLP tasks require large amounts of data, and this becomes more apparent with nuanced tasks such as anaphora and coreference resolution. To combat the prohibitive costs of creating manual gold annotated data, this paper explores two methods to automatically create datasets with coreferential annotations; direct conversion from existing datasets, and parsing using multilingual models capable of handling new and unseen languages. The paper details the current progress on those two fronts, focusing on Arabic in the case of direct conversion, as well as the challenges the efforts currently face, and our approach to overcoming these challenges.

\end{abstract}

\section{Introduction and Motivation}

Anaphora resolution and coreference resolution are Natural Language Processing (NLP) tasks that involve identifying expressions that refer to the same entity \cite{aloraini2024survey}. Anaphora resolution is the process of finding the antecedent of a given anaphor \cite{lata2021comprehensive}, while coreference resolution is the task of resolving mentions that refer to the same entity into clusters \cite{aloraini2024survey}.

Anaphora resolution is crucial in Natural Language Processing (NLP) for analyzing local linguistic context and ensuring textual cohesion by interconnecting different parts of the text \cite{abolohom2021comparative, abolohom2015hybrid, lata2021comprehensive, marques2013anaphora}. This process is essential for numerous NLP applications, including machine translation \cite{abolohom2015hybrid}, information extraction \cite{recasens2010ancora}, text summarization \cite{lata2021comprehensive}, dialogue systems \cite{abolohom2015hybrid}, and question-answering systems \cite{abolohom2021comparative}. Anaphora resolution plays a vital role in simplifying expressions and maintaining contextual connections \cite{abolohom2015hybrid}. While linguists consider anaphora an elegant way to avoid repetition, it poses a significant challenge in NLP due to the complexity of identifying the referent. Without effective anaphora resolution, texts would lack full and correct comprehension.

The availability of annotated corpora with coreferential links is vital for anaphora resolution systems \cite{abolohom2017computational}. The scarcity of such resources, especially for languages other than English, can hinder progress in the field \cite{abolohom2021comparative, abolohom2015hybrid}. Several datasets have been developed for anaphora resolution in various languages, including English, Chinese, and Arabic \cite{lata2021comprehensive}. These datasets vary based on features such as domain, annotation schemes, and types of references labeled. These variations often lead to annotation inconsistencies, evaluation challenges, and domain limitations \cite{vzabokrtsky2023findings, aloraini2024survey, nedoluzhko2021one}. This, along with the need for relatively large annotated datasets to train current state-of-the-art models, motivated us to look into automatic data annotation and harmonization.

In Section~\ref{sec:lit-rev} we present a general view of current work on anaphora and coreference resolution, covering multilingual standardization efforts, and introducing current work on Arabic. 

In Section~\ref{sec:coref-ar} we describe our initial effort of automatically generating an Arabic anaphora corpus annotated in the CorefUD style. Finally, in Section~\ref{sec:coref-ud} we describe another preliminary effort of converting the entire Universal Dependencies corpus into the CorefUD format.

\section{Literature Review}\label{sec:lit-rev}

The current state-of-the-art in anaphora and coreference resolution relies predominantly on neural network-based approaches, such as \newcite{wiseman2015learning} using neural networks to transform heuristic features, and \newcite{daume2009large} introducing a neural joint approach for mention detection and coreference resolution. The approach suggested by \newcite{daume2009large} lead to stronger models, such as the one proposed by \newcite{lee2017end} which is an end-to-end model that learns to detect mentions and resolves coreferences jointly, without relying on parse trees, improving overall performance. 

In terms of the languages that these models handle, multilingual coreference resolution is currently an active area of research, with shared tasks focusing on it \cite{vzabokrtsky2022findings, vzabokrtsky2023findings, novak2024findings}.

\subsection{Standardized Annotation Efforts}

The CorefUD project is a significant effort to standardize and harmonize coreference resolution across multiple languages \cite{nedoluzhko2021coreference, nedoluzhko2022corefud}. It addresses the challenges posed by varying data formats and annotation guidelines in existing coreference corpora by creating a unified scheme and format for coreference annotation, facilitating cross-lingual research and development in anaphora and coreference resolution. It aims to provide a wide collection of coreference data in a carefully crafted, unified format, adhering to and compatible with UD design principles.

The project includes datasets for multiple languages, facilitating the development and evaluation of multilingual coreference systems. It currently includes 17 datasets for 12 languages in its public edition. The datasets are generated by harmonizing existing corpora by converting them into a common annotation scheme. This scheme is built around mentions and clusters, where every mention is a member of one, and only one, cluster, while the cluster contains all mentions referring to the same entity \cite{nedoluzhko2021coreference, nedoluzhko2022corefud}.

Another project that aims to standardize coreference annotation is the Universal Anaphora initiative, which seeks to expand the aspects of anaphoric interpretation that can be reliably annotated in anaphoric corpora \cite{poesio2024universal}. It proposes a markup scheme for encoding anaphoric information to facilitate the creation of a collection of corpora using the same scheme.

The project aims to cover all aspects of anaphoric information currently annotated in existing projects. It identifies some aspects as required while allowing projects to have the flexibility to annotate additional information.

In the current state of this work, we will be focusing on the CorefUD annotation scheme due to the availability of resources to the authors.

\subsection{Existing Work on Arabic Anaphora and Coreference Resolution}

Work in Arabic anaphora detection has identified many challenges posed by the language's complex morphology, variations, and limited resources \cite{aloraini2024survey, abolohom2015hybrid, abolohom2021comparative, aldawsari2023within, zitouni2005impact, beseiso2016coreference, abdul2011automatic}. Researchers have explored various techniques, including rule-based, machine learning, and hybrid approaches, to tackle different types of anaphora, such as pronominal, zero, and event anaphora \cite{abolohom2015hybrid, aloraini2024survey, aldawsari2023within}. However, the characteristics of Arabic which include morphological complexity \cite{abolohom2015hybrid, zitouni2005impact, bouzid2020generic} and dialect variations \cite{aloraini2024survey} continue to make work on Arabic anaphora and coreference resolution challenging.

Currently, there are a handful of annotated corpora for Arabic, such as QurAna \cite{sharaf2012qurana}, a corpus of the Quran, the A3C Corpus \cite{dahou2021a3c}, which contains a large dataset collected from a Saudi newspaper, and OntoNotes \cite{weischedel2011ontonotes}, which contains data from the Penn Arabic Treebank (PATB)  \cite{maamouri2004penn}. However, these corpora are either variant-specific, e.g. Quran, or are based on copyrighted texts, e.g. OntoNotes. Additionally, all existing corpora are annotated in their unique schemes that do not match any of the previously mentioned harmonization projects.

The absence of harmonized Arabic corpora prompted us to focus on Arabic as a case study for automatic anaphora annotation. The challenges associated with Arabic are not unique to the language; solutions applicable to Arabic may be transferable to other languages, especially Semitic languages. Existing annotation efforts, such as the Prague Arabic Dependency Treebank (PADT) \cite{smrz2006information}, could aid in knowledge transfer across unrelated languages with similar schemes. Additionally, Arabic being the first language
of the first author facilitates the data analysis process.

\section{CorefUD Dataset for Arabic}\label{sec:coref-ar}

Following the previous effort of transforming the English OntoNotes annotations into the CorefUD scheme, we converted the Arabic OntoNotes dataset into the same scheme as described in \newcite{nedoluzhko2022corefud}.

\subsection{Data}

For this experiment, we used OntoNotes' Arabic dataset \cite{ontonotes5}. The data set comprises 599 articles with approximately 400K tokens. Since the data is entirely from the Penn Arabic Treebank, it only contains newswire articles.

The annotations contained in OntoNotes are organized in layers, as follows:
\begin{itemize}
    \item \textit{Treebank}:
    This first layer consists of the syntactic annotations of the sentences. These annotations are the parses that are provided by the LDC for the data in the PATB part~3~- v3.1 \cite{maamouri2004penn}.

    \item \textit{Proposition}: The purpose of the propositional layer is
        to identify predicate constituents and their arguments. It serves
        to resolve the roles that different components of a sentence play.
        For example, the object of an active verb and the subject of the
        equivalent passive verb are roles that should theoretically be
        played by the same entity. Since syntactic parses fail to provide
        that level of information, OntoNotes relies on the output of The
        Proposition Bank (PropBank) \cite{bonial2014propbank}. In the
        Arabic OntoNotes dataset, 87.8\% of the tokens are annotated with
        propositional information.

    \item \textit{Word Sense}: These annotations distinguish the different
        meanings a word can have. They help improve system performances on
        tasks such as information extraction and summarization. Word sense
        annotations are provided for a subset of both verbs and nouns. In
        the Arabic OntoNotes dataset, 54.8\% of the verb and 16.8\% of the
        nouns have word sense annotations.

    \item \textit{Ontology}:
    In this layer of annotations, the various word senses from the previous layer are connected to the relevant nodes of the Omega ontology \cite{philpot2005omega}. This allows for different words that share meanings or are related to each other to be connected with the relevant relations and features.

    \item \textit{Coreference}:
    For our current purposes, this layer contains the most relevant information. The annotations in this layer connect names, nominal references, and pronouns that refer to the same entity, marking them as coreferents. Similarly, verbs and their equivalent noun phrases are also marked as coreferents. These annotations can span multiple sentences as long as they occur in the same document. Appositions are also marked in this layer of annotations. In the Arabic OntoNotes dataset, only 447 articles are annotated for coreference, making a total of 319K annotated tokens.

    \item \textit{Named Entities}: Names in this layer are categorized into one of 11 categories, including persons, countries, companies, and others. Other nouns are also annotated to indicate categories such as date, time, and monetary values among others.

\end{itemize}

The tags that are used by OntoNotes are the ones that currently appear in our converted files. The first tag that OntoNotes uses is IDENT, denoting any nominal mentions of the same entity. This is reflected in our dataset by giving the entities the same IDs, without any further elaboration on tags. The second tag in OntoNotes is APPOS, denoting appositives: nominal phrases following other nominal phrases to provide extra information, rename, or further define the first mention. The APPOS tag is combined with either the HEAD subtype, denoting the referent or initial nominal phrase, and one or more ATTRIB, denoting the attributes of the HEAD.

\subsection{Analysis}

As an initial step, following the conversion rules that were applied to the English OntoNote dataset, we obtain the current first version of the Arabic CorefUD dataset. Table~\ref{table:coref-ud-stats} presents some statistics of this dataset.

\begin{table}[h!]
\begin{tabular}{|l|l|l|}
\hline
                    & \# of tokens & \# of sentences \\ \hline
Total               & 299,400      & 9,375           \\ \hline
Part of an entity   & 37,227       & -               \\ \hline
Single token entity & 21,107       & -               \\ \hline
Nested entity       & 4,670       & -               \\ \hline
\end{tabular}
\caption{Entity annotations generated by the automatic conversion of the Arabic OntoNotes dataset into the CorefUD format.}
\label{table:coref-ud-stats}
\end{table}

As for the annotations that were produced by this conversion, Table~\ref{table:coref-ud-annotations} shows some of the relevant statistics.

\begin{table}[h!]
\centering
\begin{tabular}{|l|l|}
\hline
                    & \# of tokens \\ \hline
HEAD                & 1,748        \\ \hline
ATTRIB              & 1,788        \\ \hline
APPOS               & 1,785        \\ \hline
\end{tabular}
\caption{The statistical distribution of the different labels that were generated by the automatic conversion of the Arabic OntoNotes dataset into the CorefUD format. 
}
\label{table:coref-ud-annotations}
\end{table}

A brief error analysis of the sentences in the first two documents showed that these annotations were accurate. The first two documents contained 15 and 26 sentences respectively, with 76 and 151 annotated tokens respectively. However, this sample size is too small to arrive at a conclusive evaluation of the quality of the conversions, and as such, a more in-depth error analysis will be carried out in the next phase of this effort.

\subsubsection{Zeros}

In linguistics, especially in the context of pro-drop languages, subject pronouns can be omitted; these omitted subjects are called zero pronouns \cite{aloraini2024survey}. However, even when these pronouns are dropped, they can still be part of a coreference chain. As such, in order to identify all the coreference occurrences in a text, zeros must be identified and inserted in their appropriate locations. Additionally, not all zeros need to be part of a coreference chain, and making this distinction is another task that a coreference resolution model needs to learn.

As the current conversion approach is based on English, which does not contain zeros, the generated output does not cover this linguistic phenomenon. Fortunately, the PATB annotations included within OntoNotes contain zero nodes, and the subsequent OntoNotes coreference annotations consider the zeros. Therefore, our next step in the automatic conversion process is to make use of the existing information in order to improve the coverage of the annotated entities in our conversion output. 

\section{CorefUD Dataset for the Universal Dependencies Corpora}\label{sec:coref-ud}

In an attempt to bridge basic UD and deeper semantic representations, \newcite{schuster2016enhanced} propose adding a layer of enhanced annotations to the existing UD annotation. Additionally, \newcite{droganova2019towards} propose introducing a deep-syntactic annotation layer that can be derived semi-automatically from surface UD graphs to provide a deeper level of linguistic annotation for natural language understanding. Even though neither of these annotations are required in the UD corpora, they provide crucial information that could be geared towards generating automatic anaphora annotations.

In an attempt to kickstart this effort, we turned to CorPipe \cite{straka2022ufal, straka2023ufal, straka2024corpipe}, the winning system of the three CRAC Multilingual Coreference Resolution shared tasks \cite{vzabokrtsky2022findings, vzabokrtsky2023findings, novak2024findings}. CorPipe comes with pre-trained multilingual models that can handle new and unseen languages without the need for pretraining. As a preliminary step, we ran CorPipe24 \cite{straka2024corpipe} on the entire Universal Dependencies 2.15 corpus. Our decision to go with this version of the system was its ability to predict zero mentions without the need to introduce an external pre-processing process.

\subsection{Analysis and Next Steps}

The current Universal Dependencies 2.15 corpus contains 598 treebank files over 296 treebanks and 168 languages. Due to unforeseen issues in our setup, the CorPipe24 system was only able to parse 424 files, making up only 70\% of the files. Further analysis of the causes of this failure will be required before moving forward to the next step.

Another challenge we are anticipating is the evaluation of the quality of the parsed output since finding existing gold annotations of CorPipe's output on the UD treebanks is not guaranteed. We know that some UD treebanks, such as GUM for English, PDT for Czech, and AnCora for Catalan and Spanish contain gold standard coreference annotations. Since these annotations are already part of CorefUD, they were used in CorPipe's training data. For these datasets, we will be able to evaluate the test portions of these treebanks, but it will not be a zero-shot scenario. Other datasets have coreference annotations that are not part of the current UD release. As such, we will have to try to leverage existing gold annotations in other datasets that could be easily transformed into the appropriate format, as well as manually annotate a sample across multiple languages and language families.


Knowing the quality of our parses will allow us to confidently try and retrain the CorPipe model with more languages, in the hopes of improving its performance on new and unseen languages. Additionally, we will be able to evaluate other systems using the same gold data.

\section{Evaluation Challenges and Future Work}

The experiments detailed in this paper aim to automatically generate annotations to address challenges like small existing corpora and the high costs of manual annotations for large datasets. However, to evaluate the quality of these automatic annotations, we need some gold standard data, which is currently unavailable. As a result, we plan to manually verify a portion of these annotations across as many languages as possible. Due to time constraints, this part of the experiments will be conducted after the publication of this paper.

\section{Conclusion}

This paper presented a preliminary and ongoing effort to identify potential methods of automatically generating anaphora annotation. We are exploring an approach of directly converting existing datasets into a unified format, namely CorefUD, as well as using pre-trained multilingual models to automatically parse larger multilingual corpora that can be used to retrain and improve the existing models. Both efforts involve the manual creation of gold datasets for evaluation purposes.

The two efforts result in two datasets: 1.\ Arabic OntoNotes converted to the CorefUD format; 2.\ Universal Dependencies treebanks (in many languages) with coreference annotation added by the CorPipe multilingual model.
Since this paper is a mid-point description of the effort, current data is not in a state we feel comfortable sharing. However, all final conversion and parsing outputs will be publicly available, along with proper evaluations and in-depth error analysis.

\bibliography{coref-ud}
\bibliographystyle{acl_natbib}

\end{document}